\documentclass{article}

\usepackage{microtype}
\usepackage{graphicx}
\usepackage{subcaption}
\usepackage{booktabs} % for professional tables

\usepackage{hyperref}

\usepackage[preprint]{icml2026}

\usepackage{amsmath}
\usepackage{amssymb}
\usepackage{mathtools}
\usepackage{amsthm}

\usepackage[capitalize,noabbrev]{cleveref}

\theoremstyle{plain}

\theoremstyle{definition}

\theoremstyle{remark}

\usepackage[disable,textsize=tiny]{todonotes}
\icmltitlerunning{Diagnosing Seed Exposure for LLM Ideation}

\begin{document}

\twocolumn[
  \icmltitle{Read, Grep, and Synthesize: Diagnosing Cross-Domain \\
    Seed Exposure for LLM Research Ideation}

  \begin{icmlauthorlist}
    \icmlauthor{Yunju Choi}{yonsei}
    \icmlauthor{Min Song}{onomaai}
  \end{icmlauthorlist}

  \icmlaffiliation{yonsei}{Yonsei University, Seoul, Republic of Korea}
  \icmlaffiliation{onomaai}{OnomaAI, Seoul, Republic of Korea}

  \icmlcorrespondingauthor{Min Song}{min.song@yonsei.ac.kr}

  \icmlkeywords{LLM research ideation, cross-domain retrieval,
    rubric-based evaluation, agentic workflows}

  \vskip 0.3in
]



\printAffiliationsAndNotice{}  % no special notice (required even if empty)

\begin{abstract}
  The discovery of novel methodologies for emerging problems is a continuing
  cycle in ML, often driven by the migration of techniques across domains.
  Building on this observation, we ask whether current LLM ideation systems
  benefit from \emph{targeted} cross-domain retrieval or simply from
  exposure to diverse mechanisms. We study this question through PaperGym,
  a three-stage pipeline: (1)
  tool-augmented seed extraction via read, grep, and bash over an isolated
  paper environment, (2) cross-domain seed retrieval via paraphrasing across
  seven ML domains, and (3) method synthesis from retrieved seeds, each
  scored by rubric-based judges. Tool-augmented extraction improves
  specificity, and paraphrase-based retrieval broadens domain coverage. In
  synthesis, cross-domain retrieval receives more pairwise novelty wins than
  no-retrieval and same-domain baselines, but shows no significant difference
  from a random diverse-seed control. These findings suggest LLM ideation
  systems benefit from diverse seed exposure, but do not yet reliably
  exploit the semantic reason particular seeds were retrieved. We release the seed library, rubric prompts, and run scripts at
  \url{https://github.com/yunjoochoi/PaperGym}.
\end{abstract}

\section{Introduction}
\label{sec:intro}

Large language models are increasingly applied to research workflows
\citep{huang2023mlagentbench}: surveying literature, drafting code,
running ML experiments, and proposing experimental directions.
Among such tasks, the open-ended generation of novel research ideas
remains a demanding task: the framings that yield
strong ideas typically lie beyond direct retrieval, motivating
reformulations that view the problem through alternative lenses.

One reliable source of novelty in human research is the migration of
techniques across domain boundaries. Gradient-based adversarial
optimization, prominent in adversarial examples for vision
\citep{goodfellow2015adversarial}, resurfaced as GCG
\citep{zou2023gcg} for LLM jailbreaking. Earlier work like
SOLVENT \citep{chan2018solvent} showed that human-annotated
facet labels on research papers can support cross-domain
analogy retrieval. Whether LLMs can systematically exploit
such cross-domain migration during research ideation, beyond
open-prompted analogy generation \citep{ding2023fluid},
has not been comparatively tested against simpler
diverse-exposure alternatives.

We address this question with \textbf{PaperGym}, a
three-stage pipeline that explicitly separates
(i) interactive seed extraction from a paper environment,
(ii) cross-domain seed retrieval through paraphrasing, and
(iii) method synthesis from retrieved seeds. A tool-augmented
agent extracts grounded \emph{seeds} from candidate papers via
read, grep, and bash over an isolated sandbox. Seeds are
retrieved across seven ML domains by paraphrasing the problem
statement. The retrieved seeds are then composed into a
candidate method. This compositional pattern parallels recent
prompt synthesis \citep{liu2024autodan, xiong2025cop}. We
extend it to research ideation, with papers as the
compositional primitives. Each stage is paired with a judge
using rubrics for reusable scoring.

We focus on a single stage of the research workflow rather
than full end-to-end research automation \citep{ai-scientist,
ai-researcher, schmidgall2025agentlab}, which runs over long
horizons and complicates attribution of outcomes to specific
design choices. Closer to our setting, ResearchAgent
\citep{baek2024} targets ideation through a citation graph
and an entity-centric knowledge store of shared concepts
across papers, and SciMON \citep{wang2024scimon} retrieves
inspirations from past papers and optimizes for novelty by
iterating against literature. Unlike these 
systems, we explicitly paraphrase each
problem into seven target-domain vocabularies before
retrieval, and compose the retrieved seeds via attributed
multi-seed synthesis.

To diagnose what drives ideation quality, we evaluate four ablation
conditions on a 30-problem benchmark: no retrieval, same-domain
retrieval, cross-domain retrieval, and a deliberately uninformative
random-seed control. We investigate:

\begin{itemize}
  \item \textbf{Extraction.} Does an agent equipped with read,
        grep, and bash over a paper sandbox extract more specific
        and well-grounded seeds than direct extraction without
        these tools?

  \item \textbf{Retrieval.} Does paraphrasing the problem
        statement across seven ML domains broaden the domain
        coverage of retrieved seeds without sacrificing relevance,
        compared to retrieval without paraphrasing?

  \item \textbf{Synthesis.} Does cross-domain seed exposure yield
        more novel methods than no-retrieval and same-domain
        baselines without sacrificing validity or coherence, and
        is any resulting gain driven by seed \emph{content} or
        merely by the \emph{presence} of a diverse seed pool?
\end{itemize}

Our findings are asymmetric: cross-domain retrieval receives more
pairwise novelty wins than no retrieval (60\% vs 40\%) and
same-domain retrieval (67\% vs 30\%, $3\%$ tie), and matches
a random-seed control (47\% vs 53\%). We take this as a diagnostic signal
that current LLM ideation systems benefit from diverse
mechanism exposure, but do not yet reliably exploit why
particular seeds were retrieved.
Section~\ref{sec:discussion} returns to this distinction.

\section{The PaperGym Pipeline}
\label{sec:pipeline}

\begin{figure*}[t]
  \centering
  \includegraphics[width=\linewidth]{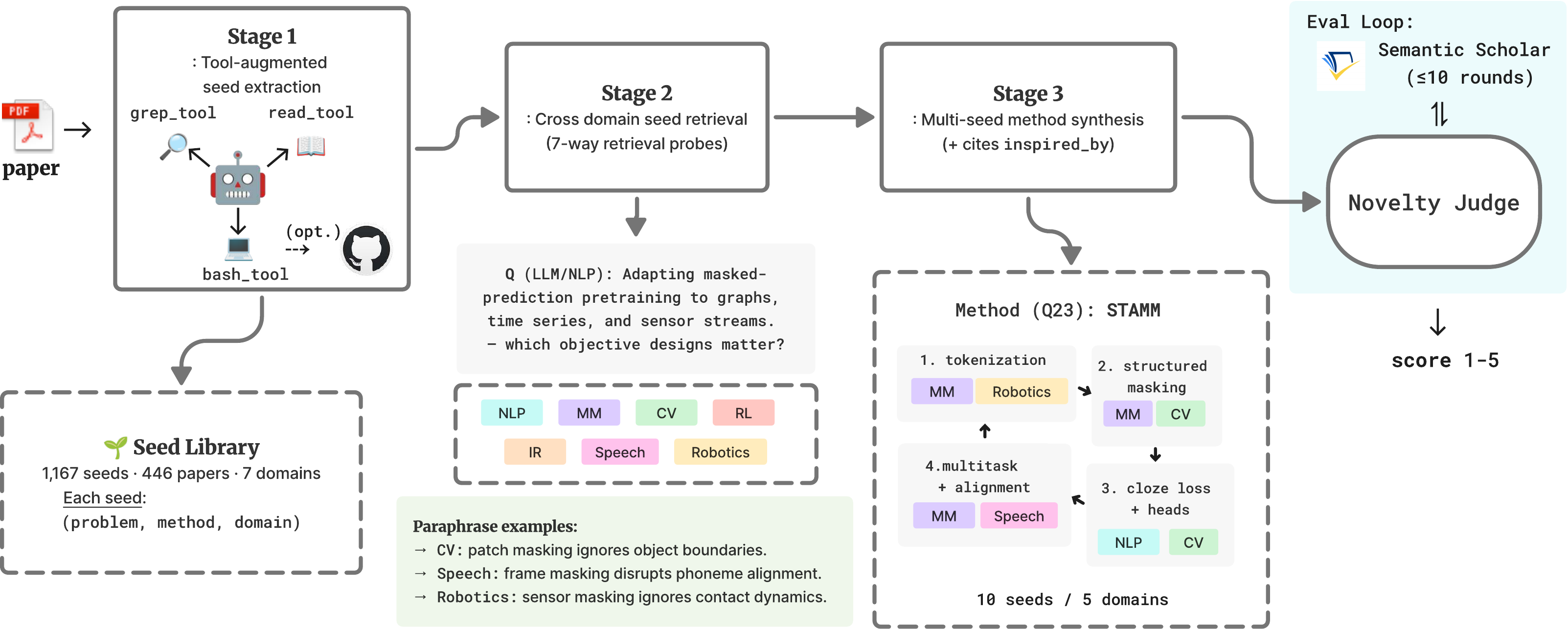}
  \caption{The PaperGym three-stage pipeline.
    \emph{Stage 1} extracts grounded seeds from candidate papers
    using an LLM agent with read, grep, and bash over an isolated
    sandbox. \emph{Stages 2--3} paraphrase the original problem
    across seven ML domains, retrieve top-$k$ seeds per paraphrase
    from the resulting library, and synthesize a candidate method
    with attribution to the contributing seeds.}
  \label{fig:pipeline}
\end{figure*}

\subsection{Tool-Augmented Seed Extraction}
\label{sec:pipe-extract}

The first operation extracts \emph{seeds} from candidate papers:
short summaries of a method's underlying problem and its
proposed mechanism. Each seed records a
\texttt{(problem, method, domain)} triple anchored to a source
paper.

Extraction runs as an LLM agent equipped with read, grep, and
bash tools. The agent operates in an isolated sandbox containing
the paper's full text in markdown form. If the paper provides
an official code repository, the agent clones it dynamically for
inspection during extraction. The agent first reads the paper
end-to-end, identifies one to three distinct contributions,
drafts candidate seeds, and then verifies each seed by locating
its central claims (mechanism name, key design choice, or
quantitative finding) via grep or additional reading;
unverified claims are revised or dropped. This re-reading step distinguishes
tool-augmented extraction from \emph{direct extraction}, where
the agent receives the entire paper as prompt context and emits
seeds in a single pass without tool calls.

\subsection{Cross-Domain Seed Retrieval}
\label{sec:pipe-retrieve}

Given a research problem, the second operation retrieves seeds
from the library across the seven ML domains (enumerated in
Section~\ref{sec:study-setup}). The original problem statement
is paraphrased into six domain-specific restatements covering all
ML domains except its natural domain. The natural domain is
probed with the raw problem statement directly. Each restatement
preserves the underlying problem while recasting it in that
domain's native research vocabulary and concept structure
(Figure~\ref{fig:pipeline} shows the Q23 case).

Each restatement (and the raw problem for the natural domain) is
embedded and used to retrieve, by problem statement similarity,
the top-$k$ seeds from the library. Concretely, with embedding
function $\phi$, paraphrase $\rho_d(q)$ for domain $d$
(setting $\rho_{d^\star}(q)\!=\!q$ on the natural domain
$d^\star$), and the merged library $\mathcal{L}$, the seed
pool retrieved for query $q$ is
\begin{equation}
\mathcal{S}(q) \;=\; \bigcup_{d \in \mathcal{D}}
\mathrm{Top}\text{-}k\,\big(\phi(\rho_d(q)),\ \mathcal{L}\big),
\label{eq:retrieve}
\end{equation}
with $|\mathcal{D}|=7$, $k=3$, and paper-id deduplication,
so $|\mathcal{S}(q)| \le 21$. We choose $k=3$ in
Eq.~\ref{eq:retrieve} to match the single-probe baseline's
$21$-seed budget. Retrieval is
global across all domain shards, so the paraphrase
vocabulary itself biases the top-$k$ toward seeds from the
target domain rather than partitioning by shard.

\subsection{Multi-Seed Method Synthesis}
\label{sec:pipe-synthesize}

The third operation composes the retrieved seeds into a candidate
method for the original research problem. The synthesizer also
produces a rationale articulating the methodological choices and
an \texttt{inspired\_by} list naming the seeds the method draws
on, each annotated with the \emph{borrowed aspect} taken from
that seed. This structure encourages coherently composed,
source-grounded synthesis. Each seed is presented with a
\emph{lens}: the text used to retrieve it, indicating
why the seed was considered.

\section{Empirical Study}
\label{sec:study}

\subsection{Setup}
\label{sec:study-setup}

\paragraph{Benchmark.} We evaluate PaperGym on a benchmark of $30$
research problem statements across seven ML domains: LLM/NLP,
multimodal, computer vision, reinforcement learning,
IR/recommendation, speech, and robotics. Each problem follows the
form ``\textit{[gap]; how can we [research question]?}'',
combining a stated bottleneck with a directional research
question.

\paragraph{Seed library.} The retrieval library contains $1{,}167$
seeds extracted by the tool-augmented agent from $446$ 
conference papers ($2017$--$2025$) distributed across the seven
domains. Venue allowlist and per-domain budget in
Appendix~\ref{app:repro}. For a paired Stage 1 ablation
(Section~\ref{sec:study-stage1}), we additionally extract seeds
from $30$ papers using direct prompting without tools. These
direct-extraction seeds are scored only for Stage 1 quality and
do not enter the retrieval library.

\paragraph{Models.} GPT-5 generates all outputs. Claude Sonnet
4.6 judges. Different model families mitigate self-enhancement
bias \citep{zheng2023judging}, and pairwise positions are
randomized per call to mitigate the position bias documented
for LLM judges \citep{wang2024fairevaluators}.

\paragraph{Rubrics.} Each stage is scored by rubric-based
rating judges \citep{liu2023geval}; Stage 3 additionally
uses pairwise judges with positions randomized per call.
Stage 1 rates seed-level \emph{specificity} and
\emph{grounding} ($1$--$5$); Stage 2 reports
\emph{domain coverage} ($1$--$7$) and \emph{relevance}
(naive and lens-aware, $1$--$5$); Stage 3 rates
method-level \emph{novelty} and \emph{validity} ($1$--$5$),
plus pairwise novelty/validity/coherence and auxiliary
\emph{seed incorporation} ($1$--$3$) and \emph{attributed
domain coverage}. Full rubric definitions appear in
Appendix~\ref{app:rubrics}.

\paragraph{Ablation.} For Stage 3 we compare four ablations:
\textbf{A} (no retrieval, problem only); \textbf{B} (same-domain
retrieval, top $21$ seeds from the problem's domain);
\textbf{C} (cross-domain retrieval, top-$k$ seeds aggregated
across the seven paraphrase domains); and \textbf{D} (random-seed
control, $21$ seeds drawn uniformly from the library). A, B,
and C progressively broaden the retrieval scope 
while keeping the seed budget fixed at $21$ from B onward. D tests
whether the gains observed under C depend on seed content or merely
on seed presence.

\subsection{Stage 1: Tool Augmentation Improves Seed Specificity}
\label{sec:study-stage1}

We compare \emph{tool-augmented extraction} (Section~\ref{sec:pipe-extract})
against a \emph{direct extraction} baseline that emits seeds in
a single prompt without tool calls, on $30$ paired papers. Each
paper's seed pool is scored along the Stage 1 rubric and
aggregated to a paper-level mean.

\paragraph{Specificity.} Tool augmentation lifts mean seed
specificity from $4.22 \pm 0.57$ (direct extraction) to
$4.76 \pm 0.22$ (tool-augmented extraction), a $+0.54$ absolute
gain with markedly lower spread. This is consistent with
interactive paper-environment exploration (navigating code, and
re-reading sections of the paper to verify each seed's grounding,
much as a human would) producing more concretely
specified seeds than direct prompting alone.

\paragraph{Grounding.} Both conditions score similarly on
grounding ($4.82$ in both settings), indicating that the gain in
specificity does not come at the cost of factual fidelity to the
source paper.

\paragraph{Negative control.} To verify that the grounding
rubric responds to actual fidelity rather than reflecting judge
leniency, we re-ran the grounding judge on a shuffled-paper
control where each seed is paired with a random unrelated paper.
Grounding collapses to $1.00$ in $100\%$ of cases ($n{=}85$ for
direct extraction, $n{=}82$ for tool-augmented extraction; mean
drop $\approx 3.82$). High grounding scores therefore reflect
real paper-seed alignment, not indiscriminate high marks.

\subsection{Stage 2: Paraphrasing Achieves Complete Cross-Domain Coverage}
\label{sec:study-stage2}

We compare two retrieval modes that share a $21$-seed budget:
paraphrase mode (top-$3$ per domain) and a single-probe
baseline (top-$21$ from the original problem).

\paragraph{Coverage.} Paraphrase mode achieves domain coverage
of $7.00 \pm 0.00$: every one of the $30$ problems retrieves
seeds spanning all seven domains. Single-probe retrieval
reaches $5.03 \pm 1.22$. Paraphrasing improves coverage
in $28/30$ problems; the two non-improvements already
span all seven domains without paraphrasing.

\paragraph{Relevance.} On the $1$--$5$ relevance rubric
(Appendix~\ref{app:rubrics}), paraphrase mode scores naive
$1.73$ and lens-aware $1.91$; the single-probe baseline scores
$2.16$ (lens $=$ naive by construction). The lens-aware
lift shows the paraphrase frame helps the judge recognize
cross-domain analogical fit that surface vocabulary alone
obscures. Paraphrase mode still trades per-seed relevance for full
coverage relative to single-probe. We next ask whether this
broader substrate translates into better synthesis.

\subsection{Stage 3: Diverse Exposure Helps, Targeting Remains Unresolved}
\label{sec:study-stage3}

We measure the quality of synthesized methods across conditions
A--D (defined in Section~\ref{sec:study-setup}). Pairwise novelty
is the primary axis; rating-scale scores, validity, and
attributed domain coverage serve as supporting evidence. The
novelty judge follows a ReAct loop \citep{yao2023react} that
augments its rating with Semantic Scholar \citep{kinney2023s2}
searches (up to 10 rounds), letting it check the synthesized
method against published prior work rather than rely on
parametric memory alone.

\begin{figure}[h]
  \centering
  \includegraphics[width=\linewidth]{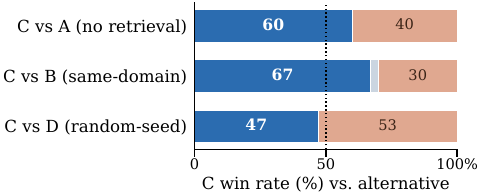}
  \caption{Stage 3 pairwise novelty win rates ($n=30$;
    positions randomized). C trends above A and B but
    is on par with D.}
  \label{fig:pairwise-novelty}
\end{figure}

\paragraph{Pairwise novelty.} Cross-domain seed exposure (C)
receives more novelty wins than both no-retrieval (A) and
same-domain retrieval (B); relative to the random diverse-seed
control (D), however, C neither wins nor loses at this sample
size (Figure~\ref{fig:pairwise-novelty}).\footnote{Two-sided
binomial $p$-values: $0.36$ for C vs A, $0.06$ for C vs B,
$0.86$ for C vs D ($n{=}30$).} Per-condition novelty averages
mirror this picture (A: $3.97$, B: $4.07$, C and D tied at
$4.13$).

\paragraph{Validity.} Per-condition validity averages cluster
near ceiling ($4.97$--$5.00$ across all conditions) and provide
little discriminative signal. The pairwise validity breakdown
appears in Appendix~\ref{app:aggregates}.

\paragraph{Coherence.} We additionally ran a pairwise coherence
judge measuring whether each method's mechanism is internally
consistent and implementable as written. C wins $17/30$ against
D (binomial $p \approx 0.58$).
Random-seed methods are thus not detectably less coherent than
cross-domain ones; per-condition single-pass coherence shows
the same C-vs-D parity (Table~\ref{tab:per-cond}).

\paragraph{Attribution integration.}
Both C and D surface seeds from multiple domains, but
cross-domain attributions are more reliably integrated: only
$1/30$ problems under C contain a seed attributed yet not
incorporated, against $8/30$ under D (paired McNemar's exact
small-sample test Eq.~\ref{eq:mcnemar-exact}, with $\chi^2$
form Eq.~\ref{eq:mcnemar}; $p \approx 0.016$). Per-condition means and the discordant-pair
breakdown appear in Appendix~\ref{app:aggregates}.

\section{Discussion}
\label{sec:discussion}
Our results support a diagnostic conclusion rather than a
simple claim that cross-domain retrieval works: condition C
receives more novelty wins than both no retrieval and
same-domain retrieval, consistent with broader seed exposure
helping synthesis, but these gains disappear against the random
diverse-seed control. The parity with D points to a concrete
catalyst regime for current LLM ideation systems: they appear to
benefit from being handed a diverse set of mechanisms, whether
targeted or random, yet do not reliably convert the semantic
reason those seeds were retrieved into higher judged novelty or
coherence. The single C-over-D effect we detect is attribution integration
($1/30$ vs $8/30$ problems with abandoned citations, $p \approx 0.016$):
evidence of some sensitivity to retrieval rationale, but not enough
to produce a clear downstream advantage on our main judged
outcomes. PaperGym therefore contributes a reproducible
evaluation protocol for diagnosing when cross-domain seed
exposure helps, and where targeted retrieval has yet to deliver.

\section{Limitations and Future Work}
\label{sec:limits}

The benchmark uses $30$ problems judged by a single model
family (Sonnet 4.6); broader benchmarks, human-expert
evaluation \citep{si2024ideas}, and repeated random-pool
resampling would sharpen the conclusions. Future directions:
a rubric-rewarded RL gym for Stage 3 that trains policies to
exploit retrieval content rather than mere presence, and a
self-extending seed library.

\section*{Software and Data}

The complete seed library, rubric prompts, conversation
transcripts, evaluation outputs, and run scripts (Stages 1--3
and the Q23 walkthrough) are released at
\url{https://github.com/yunjoochoi/PaperGym}.

\section*{Impact Statement}

This work studies LLM-driven research ideation, which can
accelerate research output but introduces concerns about idea
attribution. PaperGym addresses this by recording per-seed
provenance in every synthesized method, making the source of
each borrowed mechanism inspectable. Broader risks (displacing
human ideation, concentrating research direction, inheriting
biases from the seed library) warrant attention in future
deployments.


\bibliography{bib/refs}
\bibliographystyle{icml2026}

%%%%%%%%%%%%%%%%%%%%%%%%%%%%%%%%%%%%%%%%%%%%%%%%%%%%%%%%%%%%%%%%%%%%%%%%%%%%%%%
%%%%%%%%%%%%%%%%%%%%%%%%%%%%%%%%%%%%%%%%%%%%%%%%%%%%%%%%%%%%%%%%%%%%%%%%%%%%%%%
% APPENDIX
%%%%%%%%%%%%%%%%%%%%%%%%%%%%%%%%%%%%%%%%%%%%%%%%%%%%%%%%%%%%%%%%%%%%%%%%%%%%%%%
%%%%%%%%%%%%%%%%%%%%%%%%%%%%%%%%%%%%%%%%%%%%%%%%%%%%%%%%%%%%%%%%%%%%%%%%%%%%%%%
\newpage
\appendix
\onecolumn
\section{Q23 Pipeline Walkthrough: STAMM}
\label{app:walkthrough}

This appendix traces a single condition C run of the standard
pipeline (Section~\ref{sec:pipeline}) on Q23 (cross-modal
masked pretraining): paraphrase the problem across six
non-natural ML domains, retrieve $21$ seeds via seven probes
(top-3 each) against the merged library, and synthesize a
candidate method with cross-domain attributions.
We compare the synthesized method against the no-retrieval
baseline (condition A) below; the C-vs-D comparison on Q23
appears in Appendix~\ref{app:cvd}, and the loop-based
ablation across all $30$ problems appears in
Appendix~\ref{app:loop}.

\paragraph{Problem statement.}
\begin{quote}
\textit{``Masked-prediction pretraining transformed language
and vision; how can we adapt the same principle to other data
types like graphs, time series, or sensor streams, and what
objective design choices matter?''}
\end{quote}

\paragraph{Pipeline summary.}
A single condition C synthesis on Q23 (paraphrase
$\rightarrow$ retrieve $21$ seeds via seven probes
$\rightarrow$ synthesize) produced \emph{STAMM} in approximately
$74$\,s of GPT-5 wall-clock at \$$0.07$ of API cost
(paraphraser $29$\,s, synthesizer $45$\,s; standard pricing).
The novelty judge (Section~\ref{sec:study-stage3})
subsequently rated STAMM at $4/5$ on novelty. We use the
walkthrough below as an illustrative trace of how the
C-condition pipeline composes ten attributed seeds into a
concrete method.

\paragraph{Synthesized method.}
\emph{STAMM: Structure- and Time-Aware Masked Modeling}
is a four-stage pretraining recipe that adapts masked
prediction to graphs, time series, and multi-rate sensor
streams:

\textbf{Stage 1: Modality-specific tokenization.} Lightweight
adapters turn each modality into sequences with explicit
structural or temporal indices. A graph adapter encodes nodes
and edges as tokens with type and positional codes over the
graph topology. A time-series adapter emits event tokens with
timestamps and optional segment IDs. A fusion adapter for
asynchronous sensors initializes a continuous-time latent via
a Neural CDE and samples tokens at observed times, so a single
shared transformer can attend over irregular, multi-sensor
events. This Stage~1 CDE latent serves as the continuous-time
context over which Stage~3's cloze loss is computed for
irregular sensor streams.

\textbf{Stage 2: Structure- and causality-aware masking.} For
graphs, mask nodes, edges, and small motifs as whole segments
with synchronized edge-node masking. For time series, combine
span masking with a prefix-style objective: bidirectional
within a prefix window, forecasted thereafter. For multi-rate
sensors, interleave synchronized masks (drop aligned timestamps
across channels) and channel-specific spans. Mask spans are
sampled at multiple granularities and snapped to detected
boundaries to avoid pathological holes.

\textbf{Stage 3: Energy-based cloze with decoupled heads.}
Adopt an energy-based cloze loss trained with conditional
noise-contrastive estimation: a learned, modality-aware noise
process proposes replacements for masked tokens, and the model
discriminates real from noised values, links, or attributes.
Decoupled high-capacity token heads (MLP for continuous values;
classification for discrete attributes/links) prevent the
backbone from overfitting low-level details. Auxiliary
objectives align with each structure: link prediction and
motif classification for graphs; temporal order recovery and
derivative matching for time series; cross-channel contrastive
agreement for sensors.

\textbf{Stage 4: Alignment, curriculum, and curation priors.}
A monotonic-alignment regularizer encourages attention to
progress roughly with time for streaming data. Cross-stream
attention links let one decoder pathway attend to another
stream's token-level states, tightening attribution across
heterogeneous sensors. STAMM is trained in a multitask,
instruction-style schema mixing imputation, forecasting, link
prediction, and segmentation. Dataset sampling follows
compute-aware filtering: each pool gets a utility and
repetition half-life, and steps are allocated to maximize
expected utility under a fixed budget; mask rates and span
lengths anneal short-to-long.

\paragraph{Attributed seeds.}
STAMM cites ten seed attributions across five domains
(Table~\ref{tab:q23-cites}). Multimodal contributes the
shared-backbone scaffolding and curation/instruction priors;
CV contributes the prefix-style hybrid attention and
decoupled prediction heads; LLM/NLP contributes the
energy-based cloze objective; Robotics contributes the Neural
CDE for asynchronous sensors; Speech contributes the
monotonic alignment and cross-stream attention.

\begin{table}[h]
\centering
\caption{Seeds attributed in STAMM under condition C, with
\texttt{borrowed\_aspect} mapped to the corresponding STAMM
component. Two source papers (ViLT, AIM) each contribute two
distinct seeds; their attributions are merged into a single
row, so the table's eight rows cover ten attributions.}
\label{tab:q23-cites}
\begin{small}
\begin{tabular}{p{1.6cm}p{5.5cm}p{8.4cm}}
\toprule
Domain & Source paper & Borrowed aspect (use in STAMM) \\
\midrule
Multimodal & ViLT (\texttt{arXiv:2102.03334}) &
Shallow modality adapters feeding a single shared transformer
$\rightarrow$ Stage~1's tokenization; whole-segment masking
analogous to whole-word masking $\rightarrow$ Stage~2's graph
segment masks. \\
\addlinespace
Multimodal & OFA (\texttt{arXiv:2202.03052}) &
Multitask instruction-style pretraining schema $\rightarrow$
Stage~4's mixed-task training. \\
\addlinespace
Multimodal & Scaling Laws for Data Filtering (\texttt{arXiv:2404.07177}) &
Compute-aware data filtering with utility and repetition
half-life $\rightarrow$ Stage~4's curation priors. \\
\addlinespace
CV & AIM (\texttt{arXiv:2401.08541}) &
PrefixLM-style attention mixing bidirectional context with
causal prediction $\rightarrow$ Stage~2's prefix objective;
decoupled token-level prediction head $\rightarrow$ Stage~3's
heads. \\
\addlinespace
LLM/NLP & Energy-Based Cloze (\texttt{arXiv:2012.08561}) &
Energy-based cloze objective with conditional NCE
$\rightarrow$ Stage~3's prediction loss. \\
\addlinespace
Robotics & Multiscale Sensor Fusion with Neural CDE (\texttt{arXiv:2203.08715}) &
Neural CDE latent for asynchronous multi-rate sensor streams
$\rightarrow$ Stage~1's sensor adapter. \\
\addlinespace
Speech & Modality Gap Empirical Study (\texttt{arXiv:2510.12116}) &
Monotonic alignment prior for time-indexed token sequences
$\rightarrow$ Stage~4's alignment regularizer. \\
\addlinespace
Speech & DNCASR (\texttt{arXiv:2506.01916}) &
Cross-attention linkage to align token-level states across
streams $\rightarrow$ Stage~4's cross-stream links. \\
\bottomrule
\end{tabular}
\end{small}
\end{table}

\paragraph{Cross-domain composition.}
The walkthrough's most distinctive cross-domain bridge is
the Neural CDE, attributed in our library to a Robotics
seed (Multiscale Sensor Fusion with Neural CDEs), used as
the continuous-time tokenization over which Stage~3's
energy-based cloze loss operates for irregular sensor
streams. Each individual borrowed mechanism has a published
precedent, but the rubric's Semantic Scholar search did not
surface an end-to-end clone of this particular combination.
We note that continuous-time latents (latent ODE/CDE
families) have an active line of work for irregular
time-series imputation that the rubric's search did not
surface, so the novelty claim here is relative to the prior
work that the rubric surfaced rather than a global one.

\paragraph{Comparison without retrieval (condition A).}
Running Q23 through condition A (no retrieval, problem
statement only) yields \emph{GMRP} (Generalized Masked
Reconstruction Pretraining), an in-domain method combining
Graph Transformers with Laplacian and random-walk encodings, a
Temporal Transformer/TCN hybrid for time series, and a
cross-sensor transformer for asynchronous events; multi-granular
masking (node attribute / edge / subgraph; span / spectral
patch; channel-drop) feeds a modality-aware decoder with
mixture-density and spectral losses. All cited mechanisms are
standard, well-defined components from existing
masked-pretraining literature; no cross-domain transfer occurs.

\paragraph{Pairwise assessments (STAMM vs A).}
Under the same pairwise rubrics as
Section~\ref{sec:study-stage3} (Sonnet 4.6 judge with
position randomization), the assessments are mixed
(Table~\ref{tab:q23-c-vs-a}): STAMM is preferred for novelty,
GMRP is preferred for validity and coherence.

\begin{table}[h]
\centering
\caption{Q23 pairwise assessments: STAMM (condition C) versus the
no-retrieval baseline GMRP (condition A), under the same
rubrics used in the main benchmark.}
\label{tab:q23-c-vs-a}
\begin{small}
\begin{tabular}{p{1.7cm}p{1.6cm}p{12.5cm}}
\toprule
Axis & Winner & Judge's pivot reason \\
\midrule
Novelty & STAMM (C) &
``Energy-based cloze loss trained with conditional
noise-contrastive estimation for continuous masked targets,
combined with the Neural CDE-based fusion adapter for
asynchronous multi-rate sensors and the prefix-style hybrid
bidirectional/causal attention objective.'' \\
\addlinespace
Validity & GMRP (A) &
The judge cites GMRP's ``more systematic and comprehensive
ablation framework explicitly mapping the objective design
landscape'' of in-domain choices; STAMM's narrower focus on
the cross-domain combination is judged ``less directly tied
to the modality-specific distributional challenges.'' \\
\addlinespace
Coherence & GMRP (A) &
The judge flags STAMM's ``integration of Neural CDE outputs
into the tokenization pipeline is underspecified'' and that
its energy-based cloze loss requires ``a separate
noise/proposal model whose design and training procedure are
never specified.'' \\
\bottomrule
\end{tabular}
\end{small}
\end{table}

The novelty preference reflects STAMM's cross-domain
bridges: mechanisms attributed to Robotics (Neural CDE) and
Speech (monotonic alignment, cross-stream linkage) seeds in
our library are foreign to the prior work the rubric's
search surfaced. GMRP, by contrast, stays within in-domain
objective-design space, which the judge favored on validity
and coherence; STAMM trades that within-domain tightness
for the cross-domain combination noted above. The balance
varies problem by problem in the aggregate benchmark, and
this single trace should not be read as the universal
pattern. 

\section{Q23 C-vs-D Comparison}
\label{app:cvd}

The D-condition synthesis on Q23 (random-seed control,
$21$ seeds drawn uniformly from the library) produced
\emph{MaskSCM: Masked Pretraining for Structured and
Continuous Modalities}, a four-component recipe: (1)
modality-aware tokenization yielding semantically aligned
tokens (nodes/edges/motifs for graphs; variable-length
segments or multi-scale patches for sequences; coarse
``event'' tokens for asynchronous sensors); (2) structure-
and span-aware masking curricula; (3) prediction heads and
losses matched to each modality's physics/statistics,
including continuous-latent reconstruction when
discretization is lossy; (4) a hierarchical backbone that
propagates positional/structural encodings through depth,
with cross-modal guidance when multiple sensors are
available. The synthesizer further specifies iterative
masked decoding, prototype-dissimilarity and uncertainty
regularizers, and modality-specific backbone choices (state
space blocks, Transformers, or graph Transformers).
MaskSCM also receives novelty score $4/5$, with the
synthesizer producing it in approximately $67$\,s of GPT-5
wall-clock at \$$0.06$ of API cost (D condition skips the
paraphraser).

\paragraph{Pairwise assessments.}
Under the same pairwise rubrics as
Section~\ref{sec:study-stage3} (Sonnet 4.6 judge with position
randomization), STAMM (C synthesis from
Appendix~\ref{app:walkthrough}) and MaskSCM (D) split the
three axes (Table~\ref{tab:q23-pair}): STAMM wins novelty,
MaskSCM wins validity and coherence.

\begin{table}[h]
\centering
\caption{Q23 pairwise assessments. STAMM (C) wins novelty;
MaskSCM (D) wins validity and coherence under the same
rubrics used in the main benchmark.}
\label{tab:q23-pair}
\begin{small}
\begin{tabular}{p{1.7cm}p{1.6cm}p{12.5cm}}
\toprule
Axis & Winner & Judge's pivot reason \\
\midrule
Novelty & STAMM (C) &
STAMM's ``energy-based NCE cloze loss applied to masked
prediction across heterogeneous modalities is harder to find
in prior literature as a unified design choice''; the Neural
CDE fusion adapter combined with prefix-style hybrid masking
is also ``a more distinctive mechanistic combination.'' \\
\addlinespace
Validity & MaskSCM (D) &
The judge cites MaskSCM's ``explicit matching of objectives
to each modality's physics/statistics'' of in-domain
choices; STAMM's energy-based NCE is judged ``justified more
on efficiency grounds than on directly resolving the core
challenge.'' \\
\addlinespace
Coherence & MaskSCM (D) &
The judge flags STAMM's energy-based NCE as requiring ``a
well-defined noise distribution that is non-trivial to
specify and train stably,'' noting that ``the method does
not explain how this noise process is initialized or learned
without collapse.'' MaskSCM's coherence gaps are flagged as
``more about underspecification of regularizers.'' \\
\bottomrule
\end{tabular}
\end{small}
\end{table}

As in Appendix~\ref{app:walkthrough}, this is a single
per-problem trace; it illustrates the main-benchmark C-vs-D
pattern (Section~\ref{sec:study-stage3}) at this scale:
targeted cross-domain retrieval can secure a novelty win
without simultaneously delivering on validity and coherence.
The $30$-problem aggregate remains the reference point for
generalization.

\section{Novelty Iteration Behavior}
\label{app:loop}

This appendix wraps the standard pipeline in a novelty-guided
iteration loop, in the spirit of self-feedback frameworks such
as Self-Refine \citep{madaan2023selfrefine} and Reflexion
\citep{shinn2023reflexion}, and reports an aggregate across
all $30$ problems. Each round, the synthesizer's output is
scored by the novelty judge (Section~\ref{sec:study-stage3});
if the score is below threshold ($4$ in this benchmark), the
judge's reasoning and surfaced prior work are passed back to
the synthesizer for re-synthesis (up to $10$ rounds).

\paragraph{Aggregate.}
The iteration loop is rarely triggered: $28/30$
C-condition problems and $26/30$ D-condition problems
clear the novelty threshold ($4$) on the first synthesis
attempt (mean rounds C: $1.07$, D: $1.13$), so the loop
exits without any feedback round in the vast majority of
problems. This reflects a property of the evaluation rather
than of the synthesizer: the single-pass novelty
distribution is concentrated at score $4$ (over $24$ of $30$
problems in each of conditions C and D), so a $\geq 4$
threshold rarely admits drafts into the refinement loop.
The few problems requiring a second round reach near-perfect
scores (C: $5.0$ on $n{=}2$, D: $4.75$ on $n{=}4$), and both
conditions arrive at mean novelty score $4.1/5$. Pairwise
comparisons across all $30$ problems under the same rubrics
as Section~\ref{sec:study-stage3} show no separation between
C and D on any axis (Table~\ref{tab:loop-aggregate}).

\begin{table}[h]
\centering
\caption{Loop ablation pairwise win rates over $30$ problems.
Binomial $p$ computed via Eq.~\ref{eq:binom}. None of the
three axes shows a significant C-vs-D difference; under our
novelty rubric the loop is rarely triggered (see
\emph{Aggregate} paragraph above).}
\label{tab:loop-aggregate}
\begin{small}
\begin{tabular}{lcccc}
\toprule
Axis & C wins & D wins & Ties & Binomial $p$ \\
\midrule
Novelty & $50\%$ & $50\%$ & $0\%$ & $1.00$ \\
Validity & $40\%$ & $30\%$ & $30\%$ & $0.66$ \\
Coherence & $47\%$ & $53\%$ & $0\%$ & $0.86$ \\
\bottomrule
\end{tabular}
\end{small}
\end{table}

\section{Rubric Definitions}
\label{app:rubrics}

\paragraph{Stage 1.}
Seed-level rubrics on a $1$--$5$ scale.
\emph{Specificity} measures the density of concrete technical
entities in the seed's method text, drawn from six categories:
named algorithms, datasets/benchmarks, hyperparameter $+$
value, architecture components, equations/symbols, and
procedural steps (verb $+$ object). Anchors scale by count:
$1$ for $0$--$1$ entities, $5$ for eight or more drawn from
at least two distinct categories that include a named
algorithm, a hyperparameter value, or an equation;
procedural-verb-only does not reach $5$.
\emph{Grounding} measures whether each entity in the seed has
an equivalent in the source paper (paraphrase allowed).
Hallucinations cap at $1$, invented wording at $2$--$3$ by
count, and $5$ requires verbatim/paraphrase match plus claim
fidelity (no overstating, hedges preserved).

\paragraph{Stage 2.}
The primary measure is \emph{domain coverage}: the number of
distinct ML domains ($1$ to $7$) among the retrieved seeds.
Two auxiliary $1$--$5$ rubrics score individual seeds.
\emph{Naive relevance} sees only the research problem and the
seed's \texttt{(problem, method)} text (domain label withheld)
and scores mechanism transferability: $5$ applies as-is or
addresses the same gap from a different angle, $4$
reusable with non-trivial adaptation, $3$ thematic overlap
without an actionable lever, $2$ shared area only, $1$
unrelated. \emph{Lens-aware relevance} additionally sees the
lens text (the raw problem statement for single-probe and
same-domain slots, the domain-specific paraphrase for
cross-domain slots) and is explicitly directed to ignore lens
eloquence: a fluently written lens cannot rescue a seed whose
mechanism is genuinely unrelated. For single-probe retrieval
the lens equals the raw problem statement, so the two scores
coincide by
construction; for paraphrase mode they can diverge, and that
divergence isolates what the paraphrase contributed beyond
surface restatement.

\paragraph{Stage 3.}
\emph{Novelty} ($1$--$5$) is judged via a ReAct loop
\citep{yao2023react} with up to $10$ Semantic Scholar
queries: $1$ identical to a published technique on the same
problem, $2$ surface
variant of a parent technique, $3$ known mechanism
combination with prior joint application, $4$ combination
not surfaced as joint OR a named new component, $5$ a new
mechanism plus no direct near-clone. The retrieved seeds are
static context visible to the judge, not a separate scoring
axis, so the random-seed control (D) does not receive a
structural penalty. \emph{Validity} ($1$--$5$) asks whether the method's mechanism
plausibly engages the research problem's failure mode;
feasibility,
soundness, novelty, and empirical effectiveness are excluded. \emph{Pairwise judges}
on novelty, validity, and coherence compare two methods with
positions randomized per call (seed $42$); each axis focuses
on a distinct dimension (strongest novel element / most
direct failure-mode engagement / most serious flaw), and all
prompts ignore verbosity and break ties only when the methods
are genuinely indistinguishable.
\emph{Inspired-by incorporation} ($1$--$3$: no/partial/full)
checks whether each cited seed's
\texttt{borrowed\_aspect} is integral to the method, with
the hard test ``would removing it change the mechanism?''.
We additionally compute \emph{attributed domain coverage} as
a structural confirmatory signal: the number of distinct
domains among the seeds the synthesizer explicitly attributes
in its \texttt{inspired\_by} field. All point rubrics default
to the lower score on adjacent-anchor ties, suppressing
optimism bias.

\section{Single-Pass and Pairwise Results}
\label{app:aggregates}

This appendix reports the full per-condition averages and
C-centered pairwise assessments that support the main
benchmark (Section~\ref{sec:study-stage3}), plus
supplementary numbers not included in the main text.

\paragraph{Statistical tests.}
Pairwise win counts are summarized via two-sided binomial
tests on decisive outcomes (ties excluded). With $n$ decisive
comparisons and $w$ wins for the condition of interest,
\begin{equation}
p_{\text{two-sided}} \;=\;
\min\!\Bigg(1,\
2 \cdot \min\!\Bigg(
\sum_{i=0}^{w} \binom{n}{i} 2^{-n},\
\sum_{i=w}^{n} \binom{n}{i} 2^{-n}
\Bigg)\Bigg).
\label{eq:binom}
\end{equation}
For paired contrasts on inspired-by attributed-domain
coverage between conditions C and D, we use McNemar's test on
the discordant pair counts $b$ (problems incorporated only
under C) and $c$ (only under D). The standard $\chi^2$
statistic is
\begin{equation}
\chi^2 \;=\; \frac{(b - c)^2}{b + c},
\label{eq:mcnemar}
\end{equation}
but for small $b+c$ we report the exact two-sided binomial
McNemar
\begin{equation}
p_{\text{exact}} \;=\;
2 \cdot \sum_{i=0}^{\min(b,c)}
\binom{b+c}{i} 2^{-(b+c)},
\label{eq:mcnemar-exact}
\end{equation}
which yields the $p \approx 0.016$ cited in
Section~\ref{sec:study-stage3}.

\paragraph{Single-pass per-condition averages.}
For each condition we sample one method per problem ($n{=}30$)
and score it on the held-out novelty and validity rubrics
(Section~\ref{sec:study-stage3}); we additionally apply the
method-level specificity rubric on a $1$--$5$ scale (the
Stage 1 entity-density rubric adapted for longer ideation
method texts).
Table~\ref{tab:per-cond} reports the mean and standard
deviation per axis.

\begin{table}[h]
\centering
\caption{Single-pass per-condition averages across $n{=}30$
problems, mean $\pm$ std. Validity and method specificity
saturate near ceiling; coherence shows the C $\approx$ D
parity that mirrors the pairwise judgment in
Section~\ref{sec:study-stage3}.}
\label{tab:per-cond}
\begin{small}
\begin{tabular}{lcccc}
\toprule
Condition & Novelty & Validity & Coherence & Method spec. \\
\midrule
A (no retrieval)   & $3.97 \pm 0.41$ & $5.00 \pm 0.00$ & $3.63 \pm 0.49$ & $4.40 \pm 0.50$ \\
B (same-domain)    & $4.07 \pm 0.25$ & $4.97 \pm 0.18$ & $3.47 \pm 0.51$ & $4.40 \pm 0.50$ \\
C (cross-domain)   & $4.13 \pm 0.43$ & $4.97 \pm 0.18$ & $3.33 \pm 0.48$ & $4.27 \pm 0.45$ \\
D (random control) & $4.13 \pm 0.43$ & $4.97 \pm 0.18$ & $3.33 \pm 0.48$ & $4.40 \pm 0.50$ \\
\bottomrule
\end{tabular}
\end{small}
\end{table}

\paragraph{C-centered pairwise assessments.}
Pairwise comparisons follow the same single-pass methods and
the rubric-driven judge (Section~\ref{sec:study-stage3});
positions are randomized per call. Pairwise coherence is
reported only for C vs D; per-condition coherence covers all
four conditions in Table~\ref{tab:per-cond}.

\begin{table}[h]
\centering
\caption{Single-pass pairwise assessments on C-centered pairs,
$n{=}30$ per pair. ``Other'' refers to the second condition
in each pair (A, B, or D respectively). Bold marks the
largest count in each row (winner or majority tie). Binomial
$p$ computed via Eq.~\ref{eq:binom}. Pairwise validity ties
account for $30$--$43\%$ of comparisons, reinforcing the
ceiling pattern observed in the per-condition table.}
\label{tab:pairwise-c}
\begin{small}
\begin{tabular}{llcccc}
\toprule
Pair & Axis & C wins & Other wins & Ties & $p$ \\
\midrule
C vs A & Novelty   & $\mathbf{18}$ & $12$ & $0$  & $0.36$ \\
       & Validity  & $6$  & $\mathbf{15}$ & $9$  & $0.08$ \\
\addlinespace
C vs B & Novelty   & $\mathbf{20}$ & $9$  & $1$  & $0.06$ \\
       & Validity  & $9$  & $8$  & $\mathbf{13}$ & $1.00$ \\
\addlinespace
C vs D & Novelty   & $14$ & $\mathbf{16}$ & $0$  & $0.86$ \\
       & Validity  & $\mathbf{12}$ & $8$  & $10$ & $0.50$ \\
       & Coherence & $\mathbf{17}$ & $13$ & $0$  & $0.58$ \\
\bottomrule
\end{tabular}
\end{small}
\end{table}

\paragraph{Attributed-seed incorporation.}
A held-out judge scored each attributed seed on a $1$--$3$
scale (no / partial / full incorporation), applied to the
unique seeds attributed under C ($n{=}187$) and D
($n{=}191$). Average integration is comparable
(C $2.79$, D $2.73$). The discordant-pair breakdown for the
McNemar test in Section~\ref{sec:study-stage3} is: $7$
problems show at least one unused attribution under D but
not under C, versus $0$ in the reverse direction; the paired
test rejects equality at $p \approx 0.016$. Cross-domain
retrieval thus reduces the rate of completely unused
attributions but does not raise the typical level of seed
integration.

\section{Details}
\label{app:repro}

\paragraph{Models.} GPT-5 (\texttt{openai.gpt-5}) generates all
PaperGym outputs (seed extraction, paraphrasing, synthesis);
Claude Sonnet 4.6 (\texttt{anthropic.claude-sonnet-4-6}) serves
as the held-out judge for novelty, validity, coherence, and
attributed-seed grounding. Embeddings use
\texttt{text-embedding-3-small}.

\paragraph{Sandbox isolation.} The Stage 1 accumulator runs
each paper inside a Docker container.

\paragraph{Seed library.} The retrieval library contains
$1{,}167$ seeds from $446$ papers across the seven domains
(see Section~\ref{sec:study-setup}). Papers were sampled from
Semantic Scholar with publication year in $[2017, 2025]$ and a
venue allowlist of major ML/CV/NLP/IR/speech/robotics
conferences (NeurIPS, ICML, ICLR, CVPR, ICCV, ECCV, ACL, AAAI,
EMNLP, SIGIR, KDD, Interspeech, ICRA, IROS, CoRL, RecSys,
ICASSP). Per-domain target budgets were $107/89/77/60/60/60/47$
for LLM\_NLP/MULTIMODAL/CV/RL/IR\_REC/SPEECH/ROBOTICS
respectively; the resulting $446$ papers reflect the
post-sampling yield after Semantic Scholar filtering and
successful extraction.

\paragraph{Random seeds.} The random-seed control (condition D)
samples its $21$-seed pool at \texttt{seed=0}; the pairwise
judge's position-randomization RNG uses \texttt{seed=42}. Both
are fixed in the released scripts.

\paragraph{Cost.} An end-to-end re-run of the
released pipeline under GPT-5 generation
(\$$1.25$/\$$10$ per $1$M input/output tokens) and Claude
Sonnet 4.6 judging (\$$3$/\$$15$ per $1$M input/output tokens)
at standard pricing measures approximately \$$80$ of API cost
across $\sim$$10$\,h of wall-clock. Stage 3 single-pass
ideation (A/B/C/D $\times$ $30$ problems, with novelty rated
by a multi-round ReAct judge that issues Semantic Scholar
queries) dominates at \$$45$ / $4.9$\,h; the C-centered
pairwise novelty/validity judges add \$$3.4$ / $0.9$\,h, the
C-vs-D coherence judge \$$0.3$ / $<0.1$\,h, and the
novelty-iteration loop (Appendix~\ref{app:loop}) contributes
\$$29$ / $3.0$\,h. The Q23 single-trace walkthrough
(Appendix~\ref{app:walkthrough}) was measured directly at
$74$\,s and \$$0.07$ for the C-condition, consistent with the
per-problem extrapolation.

\paragraph{Per-probe deduplication.}
Within each retrieval probe, only the first seed per source paper
is kept before the top-$k$ cut; subsequent seeds from the same
paper are skipped. This prevents a single paper from dominating a
probe's slate when its seeds embed similarly.

\paragraph{Lens passing.}
Each retrieved seed is presented to the synthesizer alongside
a \emph{lens}: the text that was actually embedded to
retrieve the seed (raw problem statement, domain-specific
paraphrase, or a placeholder for the random-seed control).
The per-condition lens text is in
Table~\ref{tab:lens-per-condition}. For the random-seed
control (D) the lens explicitly states the seeds were
sampled uniformly, so any C-vs-D parity reflects synthesizer
use of seed content rather than ignorance of the
random-sampling procedure.

\begin{table}[h]
\centering
\caption{Lens text per condition.}
\label{tab:lens-per-condition}
\begin{small}
\begin{tabular}{lll}
\toprule
Condition & Seeds & Lens text \\
\midrule
A & none & (none) \\
B & top-$21$ same-domain & raw problem statement \\
C & top-$3 \times 7$ paraphrase domains & domain paraphrase per seed; raw problem for natural-domain slot \\
D & $21$ random from library & ``(uniform random sample of the library)'' \\
\bottomrule
\end{tabular}
\end{small}
\end{table}

\end{document}